\documentclass[sigconf]{acmart}
\AtBeginDocument{%
  }

\acmConference[Preprint]{}
\acmBooktitle{}
\acmISBN{}

\usepackage{tikz}
\usetikzlibrary{trees, positioning, shapes.geometric, fit, calc}
\usepackage{forest}
\usepackage{xcolor} 
\usepackage{multirow}
\newcommand\blfootnote[1]{%
  \begingroup
  \renewcommand\thefootnote{}\footnote{#1}%
  \addtocounter{footnote}{-1}%
  \endgroup
}

\begin{document}

\title{How Can Time Series Analysis Benefit From Multiple Modalities?\\A Survey and Outlook}
\author{Haoxin Liu\textsuperscript{\rm 1}$^\dag$, Harshavardhan Kamarthi\textsuperscript{\rm 1}, Zhiyuan Zhao\textsuperscript{\rm 1}, Shangqing Xu\textsuperscript{\rm 1}, Shiyu Wang\textsuperscript{\rm 2}, \\ Qingsong Wen\textsuperscript{\rm 3}, Tom Hartvigsen\textsuperscript{\rm 4}, Fei Wang\textsuperscript{\rm 5}, B. Aditya Prakash\textsuperscript{\rm 1}$^\dag$}
\affiliation{%
  \institution{\textsuperscript{\rm 1}Georgia Institute of Technology \hspace{0.2em}
  \textsuperscript{\rm 2}Bytedance Inc.  \hspace{0.2em}  \textsuperscript{\rm 3}Squirrel AI, USA   \hspace{0.2em} \\
  \textsuperscript{\rm 4}The University of Virginia \hspace{0.2em} \textsuperscript{\rm 5}Cornell University \hspace{0.2em} \\
  }
  \city{} 
  \state{}
  \country{}
}

\renewcommand{\shortauthors}{Liu et al.}
\newcommand{\Haoxin}[1]{\textcolor{purple}{[\textbf{Haoxin}: #1]}}
\newcommand{\xsq}[1]{\textcolor{blue}{[\textbf{Shangqing}: #1]}}

\begin{abstract}
  Time series analysis (TSA) is a longstanding research topic in the data mining community and has wide real‐world significance. Compared to "richer" modalities such as language and vision, which have recently experienced explosive development and are densely connected, the time-series modality remains relatively underexplored and isolated. We notice that many recent TSA works have formed a new research field, i.e., Multiple Modalities for TSA (MM4TSA). In general, these MM4TSA works follow a common motivation: how TSA can benefit from multiple modalities. This survey is the first to offer a comprehensive review and a detailed outlook for this emerging field. Specifically, we systematically discuss three benefits: (1) reusing foundation models of other modalities for efficient TSA, (2) multimodal extension for enhanced TSA, and (3) cross-modality interaction for advanced TSA. We further group the works by the introduced modality type, including text, images, audio, tables, and others, within each perspective. Finally, we identify the gaps with future opportunities, including the reused modalities selections, heterogeneous modality combinations, and  unseen tasks generalizations, corresponding to the three benefits. We release an up-to-date GitHub repository that includes key papers and resources.\footnote{\url{https://github.com/AdityaLab/MM4TSA}}
  \blfootnote{$^{\dag}$Correspondence to: Haoxin Liu <hliu763@gatech.edu> and B. Aditya Prakash <badityap@cc.gatech.edu>}
\end{abstract}
\maketitle
\begin{figure}[b]
    \centering
    \includegraphics[width=0.93\linewidth]{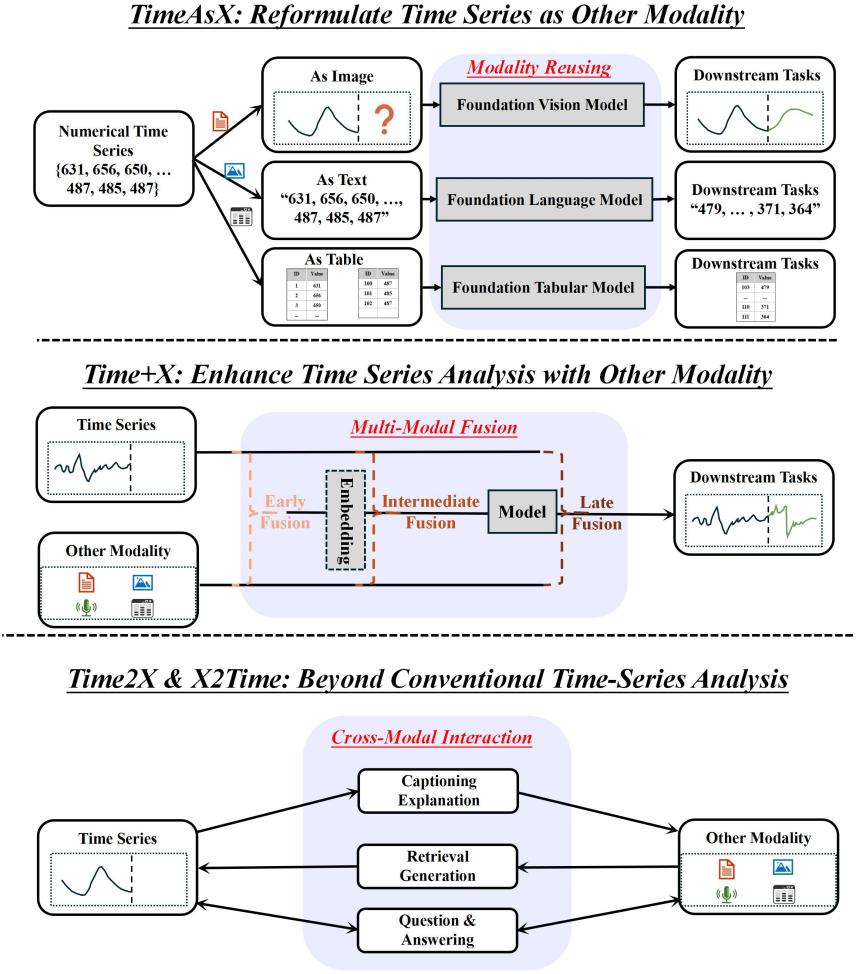}
    \caption{Illustration of three approaches of multiple modalities for TSA (MM4TSA). These three MM4TSA approaches empower TSA from modality reusing, multimodal enhancement, and cross-modal interaction, respectively.}
    \label{pic:logo}
\end{figure}
\section{Introduction}
Time series analysis (TSA) is essential across a wide range of domains, such as energy forecasting, traffic planning, and epidemic policy formulation~\cite{sezer2020financial,tabassum2021actionable,kamarthi2021doubt}. However, there is a long-standing "elephant in the room" problem: TSA research usually ignores other modalities~\cite{liu2024timemmd}. Instead, most TSA works consider only numerical series, resulting in incomplete information and non-verbalized interactions.

Recently, multiple "richer" modalities such as language and vision have recently experienced explosive development, not only as separate fields but also in multi-modal study, represented by powerful foundation multi-modal models, such as GPT and Qwen series. In contrast, the time-series modality remains relatively underexplored and isolated. We notice that many recent TSA studies have begun breaking this impasse~\cite{ming2023timellm,liu2024timemmd,liu2024picture,liu2025evaluating,merrill2024language,wang2024chattime,xie2024chatts,kong2025time}, driven by a unified high-level motivation: "How can TSA benefit from multiple modalities?"

Despite the promising prospects and rapid development of the emerging Multiple Modalities For TSA (MM4TSA) field, a systematic analysis has been notably absent. Existing studies~\cite{zhang2024large,liang2024foundation,abdullahi2025time} mainly focus on reusing large language models (LLMs) for TSA, which represents only a sub-sub-branch (one technology under one modality for one benefit), i.e., the subsection~\ref{sec:timeAsText} of this survey.

Our survey serves as \textbf{the first review of the emerging MM4TSA field} and \textbf{systematically identifies three key approaches}, as illustrated in Figure~\ref{pic:logo}: (1) TimeAsX: Reusing Foundation Models from Other Modalities for Efficient TSA; (2) Time+X: Multimodal Extensions for Enhanced TSA; and (3) Time2X and X2Time: Cross-Modality Interaction for Advanced TSA. Additionally, we \textbf{comprehensively cover multiple modalities}, including language, vision, tables, and audio, by grouping existing works according to modality type. Moreover, we introduce representative studies from specific domains, i.e., finance, medical, and spatiotemporal, to \textbf{clearly illustrate the real-world value} of these three approaches. Through this taxonomy, we \textbf{initially identify key gaps} associated with each approach: (1) which modality to reuse, (2) how to handle heterogeneous modality combinations, and (3) how to generalize to unseen tasks. We discuss potential solutions to inspire future work. Additionally, we compile a list of representative datasets and maintain an up-to-date GitHub repository of the discussed papers to \textbf{facilitate future research}.
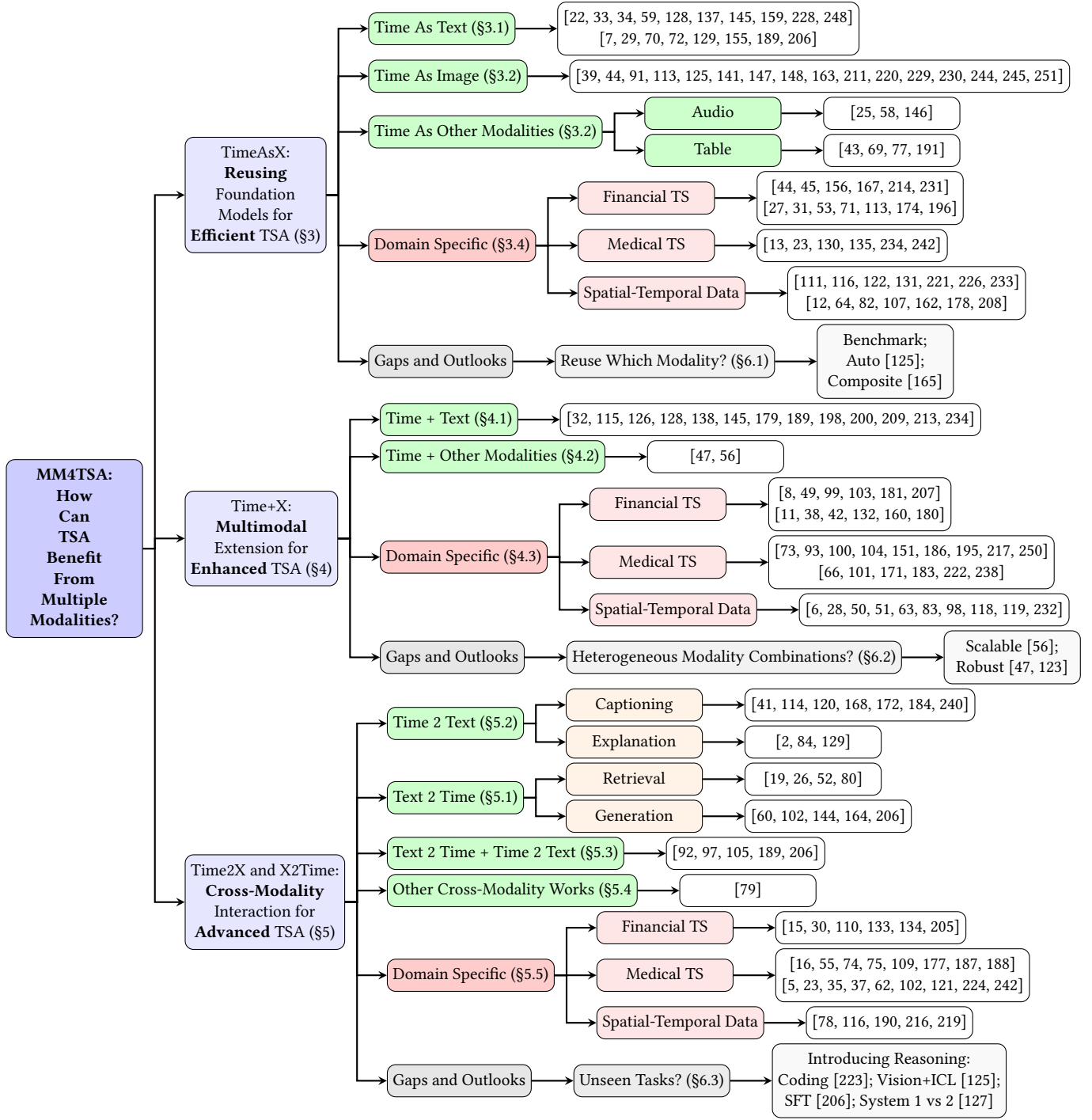
\begin{figure*}[!th]
\centering
\begin{forest}
  for tree={
    draw,
    rounded corners,
    scale=0.9,
    minimum height=0.8em,
    minimum width=8em,
    text centered,
    edge={-stealth, thick},
    edge path={
          \noexpand\path[\forestoption{edge}]
          (!u.east) -- ++(0.2cm,0) |- (.child anchor)\forestoption{edge label};
        },
    l sep=0.7cm,
    s sep=0.7mm,
    anchor=west,
    grow=east,
    align=center, 
  },
[MM4TSA:\\How \\Can\\TSA \\ Benefit \\From \\Multiple \\Modalities?, fill=blue!20, font=\bfseries, 
        [Time2X and X2Time:\\\textbf{Cross-Modality}\\Interaction for\\ \textbf{Advanced} TSA (\S\ref{sec:time2X}), fill=blue!10,
        [Gaps and Outlooks, fill=gray!20
            [Unseen Tasks?  (\S\ref{sec:crossmodalInteract}), fill=gray!15
              [
                Introducing Reasoning:\\ Coding~\cite{ye2024beyond}; Vision+ICL~\cite{liu2024picture};\\ SFT~\cite{xie2024chatts}; System 1 vs 2~\cite{liu2025evaluating} , fill=gray!5
                ]
            ]
          ]
          [Domain Specific (\S\ref{sec:time2Domain}), fill=red!20
                  [Spatial-Temporal Data, fill=red!10
                     [\cite{wang2017spatial,yang2019research,irvin2024teochat,yan2024urbanclip,li2024urbangpt}, fill=purple!0]
                  ]
                  [Medical TS, fill=red!10
                     [\cite{bleich2024automated,wan2024meit,tangelectrocardiogram,wan2024electrocardiogram,li2024frozen,hunter2008using, gatt2009data,hunter2008summarising}\\ \cite{alcaraz2023diffusion,chung2023text,lai2025diffusets,zhao2024ecg,guo2024multimodal,yuecg,liu2023biosignal,chan2024medtsllm,cosentinotowards}, fill=purple!0]
                  ]
                  [Financial TS, fill=red!10
                   [\cite{liu2022long,liu2023neural,chen2024knowledge,li2023multimodal,bhatia2024fintral,xie2024open}, fill=purple!0]
                  ]
            ]
          [ Other Cross-Modality Works (\S\ref{sec:time2Others}, fill=green!20
                [\cite{islam2024datanarrative} ,fill=purple!0]
          ]
          [Text 2 Time + Time 2 Text (\S\ref{sec:timePlusText}), fill=green!20
           [\cite{wang2024chattime,xie2024chatts,kong2025time,kim2024multi,leiitformer} ,fill=purple!0]
          ]
          [Text 2 Time (\S\ref{sec:text2Time}), fill=green!20
            [Generation, fill=orange!10
            [\cite{lai2025diffusets,merrill2024language,xie2024chatts,guverbalts,rousseau2025forging}, fill=purple!0]
            ]
            [Retrieval, fill=orange!10
            [\cite{fons2024evaluating,caitimeseriesexam,ito2024clasp,chen2025tracegroundingtimeseries}, fill=purple!0]
            ]
          ]
          [Time 2 Text (\S\ref{sec:time2Text}), fill=green!20
            [Explanation, fill=orange!10
            [\cite{aksu2024xforecast,jiang2025explainable,liu2024largeAD}, fill=purple!0]
            ]
            [Captioning, fill=orange!10
            [\cite{sharma2021t,spreafico2020neural,li2023repr2seq,zhang2023insight,dohi2024domain,lin2024decoding,trabelsi2025time}, fill=purple!0]
            ]
          ]
        ]
        [Time+X:\\\textbf{Multimodal}\\Extension for \\\textbf{Enhanced} TSA (\S\ref{sec:time+x}), fill=blue!10
        [Gaps and Outlooks, fill=gray!20
            [Heterogeneous Modality Combinations? (\S\ref{sec:modalIntegration}) , fill=gray!10
              [
                Scalable~\cite{girdhar2023imagebind};\\ Robust~\cite{ebrahimi2023lanistr,liu2025maestro} , fill=gray!5
                ]
            ]
          ]
          [Domain Specific (\S\ref{sec:timeWithDomain}), fill=red!20
              [Spatial-Temporal Data, fill=red!10
                     [\cite{anantharam2016understanding,fan2022urban,lin2022spatial,chen2024terra,feng2024citygpt,lin2023mmst,zhang2024bjtt,han2024event,jiang2024mobile,kosugi2022traffic}, fill=purple!0]
                  ]
                  [Medical TS, fill=red!10
                     [\cite{wagner2020ptb, lee2023learning, wang2024multimodal, zhu2024emerge, yang2021leverage, niu2023deep, king2023multimodal, huang2020fusion, kyono2021miracle} \\  \cite{hayat2022medfuse, yao2024addressing, tolle2025arbitrary, soenksen2022integrated, zhang2023improving, kyung2024towards}, fill=purple!0]
                  ]
                  [Financial TS, fill=red!10
                   [\cite{lee2009financial, xing2018natural,emami2023modality,kurisinkel2024text2timeseries,asgarov2023predicting,taylor2024natural}\\ \cite{liu2024multimodal,darapaneni2022stock,tavakoli2025multi,rekabsaz2017volatility,bamford2023multi,dong2024fnspid}, fill=purple!0]
                  ]
            ]
          [Time + Other Modalities (\S\ref{sec:timeWithOther}), fill=green!20
            [\cite{ebrahimi2023lanistr,girdhar2023imagebind}, fill=purple!0]
          ]
          [Time + Text (\S\ref{sec:timeWithText}), fill=green!20
           [\cite{ming2023timellm,liu2024lstprompt,williams2024context,wu2024dual,wang2024chattime,liu2025autotimes,xu2024beyond,xinlei2024from,cheng2024advancing,tao2024hierarchical,liu2024timemmd,li2025language,zhang2024dualtime}, fill=purple!0]
          ]
        ]
        [TimeAsX:\\\textbf{Reusing}\\Foundation\\Models for\\ \textbf{Efficient} TSA (\S\ref{sec:timeAsx}), fill=blue!10,
        [Gaps and Outlooks, fill=gray!20
            [Reuse Which Modality?  (\S\ref{sec:modalityreuse}), fill=gray!10
              [
                Benchmark; \\Auto~\cite{liu2024picture};\\ Composite~\cite{ruan2025vision} , fill=gray!5
              ]
            ]
          ]
          [Domain Specific (\S\ref{sec:timeAsDomain}), fill=red!20
                  [Spatial-Temporal Data, fill=red!10
                     [\cite{liu2024can,zhang2024trafficgpt,liu2024spatial,yuan2024unist,li2024urbangpt,li2025climatellm,yao2018deep} \\ \cite{li2022deep,bao2022storm,jiang2023learning,tang2024vmrnn,ren2024tpllm,xinglei2023where,hao2022leveraging}, fill=purple!0]
                  ]
                  [Medical TS, fill=red!10
                     [\cite{belyaeva2023multimodal,liu2023large,liu2024ecg,zhao2024ecg,zhang2024dualtime,chan2024medtsllm}, fill=purple!0]
                  ]
                  [Financial TS, fill=red!10
                   [\cite{xu2024quantum, sezer2019financial, pei2024stock, zeng2023pixels, dosovitskiy2020image, du2020image}\\  \cite{chen2020encoding,li2020forecasting, wang2015imaging, hu2024research,su2024mtrgl,foroutan2024deep, cheng2022financial}, fill=purple!0]
                  ]
          ]
          [Time As Other Modalities (\S\ref{sec:timeAsImage}), fill=green!20
            [Table, fill=green!20,
              [\cite{inkit2020tabular,dooley2023forecastpfn,hoo2024tabular,wang2024tabletime}, fill=purple!0]
            ]
            [Audio, fill=green!20,
              [\cite{chaohan2021voice2series,gong2022ssast,minhao2021t} ,fill=purple!0]
            ]
          ]
          [Time As Image (\S\ref{sec:timeAsImage}), fill=green!20
            [\cite{zekun2023time,luoxiao2024vitime,zhuang2024see,xu2025can,yang2024vitime,daswani2024plots,mouxiang2024visionts,dosovitskiy2020image,zeng2021deep,namura2024training,roth2022towards,zhiguang2015imaging,kim2024cafo,li2020forecasting,zhongtime,liu2024picture}, fill=purple!0]
          ]
          [Time As Text (\S\ref{sec:timeAsText}), fill=green!20
            [\cite{gruver2023large,liu2024lstprompt,zhou2023one,rasul2023lag,yuxuan2024multipatch,ching2023llm4ts,chenxi2023test,caotempo,ming2023timellm,liu2024autotimes}\\ \cite{liu2024largeAD,xie2024chatts,hu2025context,pan2024s,ansari2024chronos,wang2024chattime,chen2025large,huang2025exploiting} , fill=purple!0]
          ]
        ]
      ]
\end{forest}
\caption{A comprehensive taxonomy of MM4TSA.  \colorbox{blue!20}{Multiple Modalities For TSA (MM4TSA)} is
organized into four stages, starting with the \colorbox{blue!10}{benefit approaches} (i.e., \colorbox{blue!10}{TimeAsX, Time+X, Time2X \& X2Time)}, followed by \colorbox{green!20}{modality types} (i.e., \colorbox{green!20}{text, image, audio and table}, if available), \colorbox{red!20}{domain-specific applications} (i.e., \colorbox{red!10}{financial, medical, spatial-temporal} TSA)
and finally,  \colorbox{gray!20}{the gaps and outlooks}. For branches with extensive existing research, especially Time As Text (\S\ref{sec:timeAsText}), Time As Image (\S\ref{sec:timeAsImage}), and Time + Text (\S\ref{sec:timeWithText}), we further divide them into more detailed subcategories. }
\label{fig:taxonomy}
\end{figure*}
\section{Background and Taxonomy}
A time series, an ordered sequence of data points, can be denoted as $X_{1:T} = \{x_1, x_2, \dots, x_T\}$. Classic TSA tasks include forecasting, which predicts future values based on historical observations (e.g., flu forecasting~\cite{mathis2024evaluation}); classification, which assigns time series into discrete categories (e.g., ECG signal classification~\cite{kiranyaz2015real}); anomaly detection, which identifies abnormal patterns (e.g., financial fraud detection~\cite{ren2019time}); and imputation, which addresses missing data (e.g., completing medical records~\cite{bertsimas2021imputation}). In this survey, we exclude generalized time series such as videos~\cite{oprea2020review} or event streams~\cite{li2024mm}, since these data are inherently multimodal.  More details about background are provided in Appendix~\ref{app:back}.


The proposed taxonomy is illustrated in Figure~\ref{fig:taxonomy}. To the best of our knowledge, this is the first survey about the emerging MM4TSA field, systematically identifying three beneficial approaches, comprehensively considering various modalities, concretely introducing applications, and deeply discussing gaps and outlooks. Existing surveys~\cite{zhang2024large, liang2024foundation, abdullahi2025time} mainly focus on reusing LLMs for TSA, i.e., our Time As Text (\S\ref{sec:timeAsText}), while concurrent work~\cite{ni2025harnessing} aligns with our Time As Image (\S\ref{sec:timeAsImage}). Compared to these, our survey represents a higher level beyond "Time As X." Even within the "Time As X", we include additional modalities, such as audio and table. Regarding the choice of specific domains, we elaborate on them because they have received the most attention in existing research. However, we expect more works in other domains. We detail each branch in turn with representative works. 
\section{TimeAsX: Reusing Foundation Models of Other Modalities 
 for Efficient TSA}\label{sec:timeAsx}
 Compared with the time-series modality, language and vision modalities have richer data and deeper exploration, leading to many advanced foundation models in recent years, such as the GPT, DeepSeek, Llama, and Qwen series. These foundation models can efficiently complete various tasks in zero-shot or few-shot settings~\cite{kojima2022large,brown2020language,liu2021swin}. Thus, many recent time-series works are driven by the following motivation: Can we reuse these off-the-shelf foundation models from "rich" modalities for efficient TSA? We further classify this type of "TimeAsX" research by the reused modality into text, image, audio, and table, and introduce them one by one.

\subsection{Time As Text}\label{sec:timeAsText}

Given the success of large language models (LLMs) in a wide range of language understanding and generation tasks, recent works have explored the usage of LLMs for TSA tasks~\cite{gruver2023large,liu2024lstprompt,xinglei2023where,hao2022leveraging,zhou2023one,rasul2023lag,yuxuan2024multipatch,ching2023llm4ts,chenxi2023test,caotempo,ming2023timellm,liu2024autotimes,liu2024largeAD,xie2024chatts,hu2025context,pan2024s,ansari2024chronos,wang2024chattime,chen2025large,huang2025exploiting}. The motivation behind these works is mainly based on the fact that both language and time series have a sequence structure, and on the belief that the general ability of LLMs can be applied to TSA.

The main challenge is to align time-series data with LLMs for better understanding and activation. We  divide the related articles into three groups with representative works: (1) direct alignment without training, (2) training for alignment under an existing vocabulary, and (3) training for alignment with an expanded vocabulary.

\subsubsection{Direct Alignment without Training.}
These methods do not need any update to LLM architecture but mainly focus on how to input time-series data as text input.

\paragraph{Better Tokenization}
LLMTime~\cite{gruver2023large} simply inputs the time-series data as text to the LLM and uses the generated
output as forecasts of the time-series. To improve the ability of LLMs to properly tokenize the time series.
Another line of work involves carefully prompting the LLMs to provide context about the time-series domain and task.

\paragraph{Task-specific Prompting}
PromptCast~\cite{xue2023promptcast} provides specific text prompts to LLMs that contain information
such as domain information, time-step, task and past time-series values and prompts the LLM
to generate the next value in the time-series.
LSTPrompt~\cite{liu2024lstprompt} uses more sophisticated Chain-of-thought prompting by providing
useful sequence of steps for LLMs to reason about the time-series data, focusing on different trends and patterns for
long and short term forecasting.
Other works~\cite{manqing2024can,xinglei2023where,hao2022leveraging} have used it for tasks like anomaly detection and multi-task learning.

\subsubsection{Training for Alignment Under Existing Vocabulary}
Another line of work aligns time-series with LLMs by treating time-series as a sentence in a known vocabulary. As the dimension of both language vocabulary and time-series patches are high, most works tend to simplify the training procedure by correlating to objectives from certain time-series tasks. 
That is, the language models will be treated as time-series models by adding time-to-text transformation modules (and modules vice versa). Depending on how such transformation modules work, these works can be categorized into: 

\paragraph{Embedding Alignment}
In this line of work, time-series is aligned to existing vocabulary by training the
initial and final embedding and out layers as projections from time-series to language vocabularies or vice versa.
\citet{zhou2023one,yuxuan2024multipatch} use a frozen LLM backbone, generate patches of time-series datasets, and generate embeddings for each patch. These embeddings are then input into the
GPT2 model. The embedding layer and the output layer are then fine-tuned on the time-series data.
LLM4TS~\cite{ching2023llm4ts} additionally fine-tunes the layer normalization layers of the LLM. FreqLLM~\cite{freqlm} introduces frequency-domain adaptation.

\paragraph{Prototype Alignment}
These methods train input modules to map time-series values into fixed input embeddings (prototypes) that are closer to embedding space of pre-trained distribution~\cite{chenxi2023test}.
Time-LLM~\cite{ming2023timellm} combines text-based prompting with patch-based input.
The patches are reprogrammed to generate output embeddings via a trainable layer.
ChatTS~\cite{xie2024chatts} uses fixed attribute descriptors that capture important time-series properties like trends, noise, periodicity, and local variance.

\paragraph{Block Alignment}
~\citet{hu2025context} identifies the need to more intricately align text context with time-series when fed together as input embeddings.
They use a GNN to model the interaction between time series and text embeddings and train together with embedding and layer-norm layers of LLM when pre-training.  ~\citeauthor{qin2025bridging}  segment time series into blocks to align with langugae.

Besides, LangTime~\cite{niulangtime} propose a proximal-policy-optimization based finetuning solution.  ISTS-PLM~\cite{zhang2025unleashing} studies irregular time series.
\subsubsection{Training for Alignment with Expanded Vocabulary} 
Some methods expand LLM vocabulary to align with time-series datasets: they treat time-series data as sentences in a foreign language and adapt the LLMs toward such a language. 
These works differ in how they design the
adaptor functions to map time-series to expanded vocabulary.
Chronos~\cite{ansari2024chronos} quantizes the normalized input time-series values into discrete tokens that are fed into a language model.~\citet{wang2024chattime} similarly introduce additional tokens
for quantized values of input time-series as well as NaN or missing values.

Notably, \citeauthor{tan2024language} and \citeauthor{KDD25rethinking} rethink the actual effectiveness of reusing LLMs for TSF and various TSA tasks correspondingly.
\subsection{Time As Image}\label{sec:timeAsImage}
Reformulating time series as images for better feature perception is a natural idea, similar to how humans perceive patterns, and has received long-term research attention~\cite{zekun2023time,luoxiao2024vitime,zhuang2024see,xu2025can,yang2024vitime,daswani2024plots,mouxiang2024visionts,dosovitskiy2020image,zeng2021deep,namura2024training,roth2022towards,zhiguang2015imaging,kim2024cafo,li2020forecasting,zhongtime}. 
We look at different methods of visualizing time-series
as images for vision models in this section and provide more details Appendix \S\ref{sec:timeAsImageapp}.

Using \textit{line plots} is the most popular way to convert time series to images.
Multiple works have represented time-series as line graphs to use vision foundational models~\cite{zekun2023time,luoxiao2024vitime} like ViT.
Some works further introduce VLMs~\cite{zhuang2024see,xu2025can,shen2025multi,siru2025timevlm} and classification~\cite{daswani2024plots}.

\paragraph{Heatmaps}
Heatmaps visualize time series in a 2D space using colors to represent magnitudes.
They are specifically useful for modeling long time series~\citet{mouxiang2024visionts} and multivariate time series~\cite{zeng2021deep}
which are fed to vision~\cite{dosovitskiy2020image} and video prediction models~\cite{zeng2021deep} to generate forecasts.

\paragraph{Spectrogram}
Time series can be decomposed into the spectrum of frequencies and represented as a spectrogram.
Wavelet transforms are a popular choice of representation
for both univariate~\cite{zeng2023pixels} and multivariate~\cite{namura2024training}
tasks.

\paragraph{Other methods}
 \citet{zhiguang2015imaging} use Gramian Angular Fields (GAF)~\cite{campanharo2011duality} to represent time-series.
which visualize long and short term dependencies better.
Recurrence plots (RP)~\citet{eckmann1995recurrence} are another way to capture periodic patterns in time-series used by ~\cite{kim2024cafo}
for classification and~\cite{li2020forecasting} forecasting.
Time-VLM~\cite{siru2025timevlm}  combines information from Fourier coefficients, cosine and sine periodicity into a heatmap which is fed into a VLM encoder.

\subsection{Time As Other Modalities}\label{sec:timeAsOther}
\paragraph{Time As Audio}
Few works have tried repurposing pre-trained audio neural models for time-series analysis tasks.
Voice2Series~\cite{chaohan2021voice2series} reprograms the input time-series and feeds it to a pre-trained deep acoustic classification model~\cite{de2018neural,yang2021decentralizing}.
SSAST~\cite{gong2022ssast} instead uses filterbanks in order to align with Audio Spectrogram Transformer (AST)~\cite{gong2021ast}, an audio foundational model for classification tasks.

~\citet{minhao2021t}, similar to Wavenet, uses wavelet transform to encode patterns from multiple dominant frequencies of the time-series. They design a tree-structured network that iteratively decomposes the input signal into various frequency subbands with similar energies. This helps them provide superior performance in domains like activity recognition, brain EEG signal classification and muscular activity recognition tasks.

\paragraph{Time As Table}
Tabular foundational models perform few-shot regression or classification tasks on
tabular data.
These models can be retrofitted for time-series analysis tasks by representing time-series as tabular data with each variate a separate feature along with the time-stamps for temporal context.
For instance, \cite{inkit2020tabular}  explore the use of tabular transformers to model multivariate time series data.
\citeauthor{hoo2024tabular} introduce TabPFN \cite{hollmann2022tabpfn}, a foundational tabular model, for Time Series Forecasting (TSF). \citeauthor{hoo2024tabular} demonstrate that TabPFN, when combined with minimal featurization, can perform powerful zero-shot TSF. 
Surprisingly, this direct reusing approach has achieved leading performance (ranked 1st until March 10, 2025) on the GIFT-Eval TSF Benchmark~\cite{aksugift}. 
Further, \cite{dooley2023forecastpfn} proposes ForecastPFN, an extension of TabPFN, trained on synthetic data to perform zero-shot forecasting, thus reducing the need for extensive task-specific tuning. 
Besides, \citeauthor{wang2024tabletime} demonstrate that direct tabular reformulation is also helpful for time-series classifications.
\begin{table*}[t]
\centering
\caption{Representative Multimodal Time-Series Datasets Grouped by General, Financial, Medical and Spatial-Temporal Domains.}\label{Tab:datasets}
\resizebox{\linewidth}{!}{
\begin{tabular}{ccc}
\hline
Dataset & Modalities & Highlights \\ \hline
Time-MMD~\cite{liu2024timemmd}   &  Time+Text & 9 Domains; Real Datasets (general context); Across More than 24 Years\\ 
ChatTime~\cite{wang2024chattime} &  Time+Text & 3 Real Datasets (weather\&date)\\ 
CiK~\cite{williams2024context} &   Time+Text & 7 Domains; 71 Human-Designed Tasks; \\ 
ChatTS~\cite{xie2024chatts} &  Time+Text & Synthetic Method; 500+ Human-Labeled Samples \\ 
TSQA~\cite{kong2025time}  &  Time+Text & Multi-Task QA Format; 1.4k Human-Selected Samples  \\
\hline
FNSPID~\cite{dong2024fnspid}&  Time+Text & Large-Scale Finance News; Across 24 Years \\ 
FinBen~\cite{xie2024finben} &  Time+Text & Bilingual; 42 Sub-Datasets; 8 Tasks \\ \hline
MIMIC~\cite{johnson2016mimic,johnson2023mimic} &  Time+Text+Image+Table & Multiple Medical Tasks; Expert -Labeled Data \\
PTB-XL~\cite{wagner2020ptb} &  Time+Text & Large-Scale Expert-Labeled ECG Data\\ \hline
CityEval~\cite{feng2024citygpt}& Time+Text+Image & Multiple Urban Tasks, capable of Involving LLMs \\ 
Terra~\cite{chen2024terra} & Time+Text+Image & Worldwide Grid Data across 45 Years \\ \hline
\end{tabular}}
\end{table*}
\subsection{Domain-Specific Time-Series Works}\label{sec:timeAsDomain}
\paragraph{Financial Time Series As X}
While financial data, such as stock prices, naturally exist as numerical time series, existing methods have explored re-framing the time-series modality into alternative representations to enhance financial analysis. One of the most popular approaches is to transform financial time series into visual representations, leveraging the strengths of vision models for various financial tasks~\cite{xu2024quantum, sezer2019financial, pei2024stock, zeng2023pixels, dosovitskiy2020image, du2020image, chen2020encoding, li2020forecasting, wang2015imaging, hu2024research}. Details of these works within this line are provided in Appendix \ref{Details of Finance Time Series As Image}. With the emergence of large language models (LLMs), researchers have explored leveraging financial time series as a linguistic modality. For example, \citet{yu2023temporal} prompt LLMs with financial time series and textual financial instructions, enabling the models to predict stock trend changes. 
More works~\cite{xie2023wall, zhang2024multimodal, gan2024mme, xie2024open, gruver2023large, jin2023time, liu2024lstprompt} within this line are detailed in Appendix \ref{Details of Finance Time Series As Text}.

Beyond reframing financial time series as vision and linguistic modalities, alternative modality representations have received less attention. Possibly the only notable exploration include \citet{su2024mtrgl}, who introduces MTRGL, a framework that transforms financial time series into temporal graphs that model temporal correlations, such as relationships between stocks. 
Two more relevant works~\cite{foroutan2024deep, cheng2022financial} within this line are provided in Appendix \ref{Details of Finance Time Series As Graphs}.

\paragraph{Medical Time Series As X}
Some studies leverage pre-trained LLMs by reformulating medical time-series as tokenized sequences, prompting, or text-aligned embeddings \cite{belyaeva2023multimodal,liu2023large,liu2024ecg,zhao2024ecg,zhang2024dualtime}. Later, with the development of multimodal LLMs, MedTsLLM \citet{chan2024medtsllm} further demonstrated superiority by visually modeling medical time series, especially high-dimensional medical signals. Since medical time series are usually used with textual records, more details are discussed in \S~\ref{Med:1} and \S~\ref{Med:2}.

\paragraph{Spatial-Temporal Data As X}
A majority of spatial-temporal works aim to transform time series into textual prompts and thus introduce large language models (LLMs) as a powerful forecaster \cite{liu2024can,zhang2024trafficgpt,liu2024spatial,yuan2024unist,li2024urbangpt}. For example, UrbanGPT \cite{li2024urbangpt} proposes to convert time series to textual sequences and fine-tune an LLM with instructions containing decomposed spatial-temporal descriptions plus brief task descriptions. Details of works within this line are provided in Appendix \ref{Details of Spatial-Temporal As Text}. Besides, some also propose to convert spatial-temporal data to images \cite{yao2018deep,li2022deep,bao2022storm,jiang2023learning,tang2024vmrnn}. Specifically, they view the input data as a sequence of images corresponding to each time frame, then use vision models to encode each image to obtain stronger spatial understanding. Details of works within this line are provided in Appendix \ref{Details of Spatial-Temporal As Image}.

\section{Time+X: Multimodal Extension for Enhanced TSA}\label{sec:time+x}
Human experts typically complete TSA tasks by integrating multiple modalities, especially by combining numerical and textual data. For instance, epidemiologists combine influenza infections data with textual domain knowledge, policies, and reports to predict future trends. However, most TSA models~\cite{vaswani2017attention,liu2023itransformer,zhang2022crossformer,liu2022non,zhou2022film,kitaev2020reformer,zhou2021informer,nie2022time,liu2024time} are unimodal, using only numerical series. Extending unimodal TSA to multimodal TSA, especially through the integration of text, is an emerging topic~\cite{ming2023timellm,liu2024lstprompt,wei2022chain,williams2024context,jin2023time,wu2024dual,wang2024chattime,liu2025autotimes,xu2024beyond,dong2024fnspid,liu2024time,xinlei2024from,cheng2024advancing,tao2024hierarchical,liu2024timemmd,li2025language,zhang2024dualtime}. We first introduce multimodal TSA for general domains, starting with Time+Text and then Time+Other Modalities, followed by specific domains, using representative works. We summarize some representative multimodal time series datasets in Table~\ref{Tab:datasets}.
\begin{figure*}[t] 
\centering
  \includegraphics[width=0.8\linewidth]{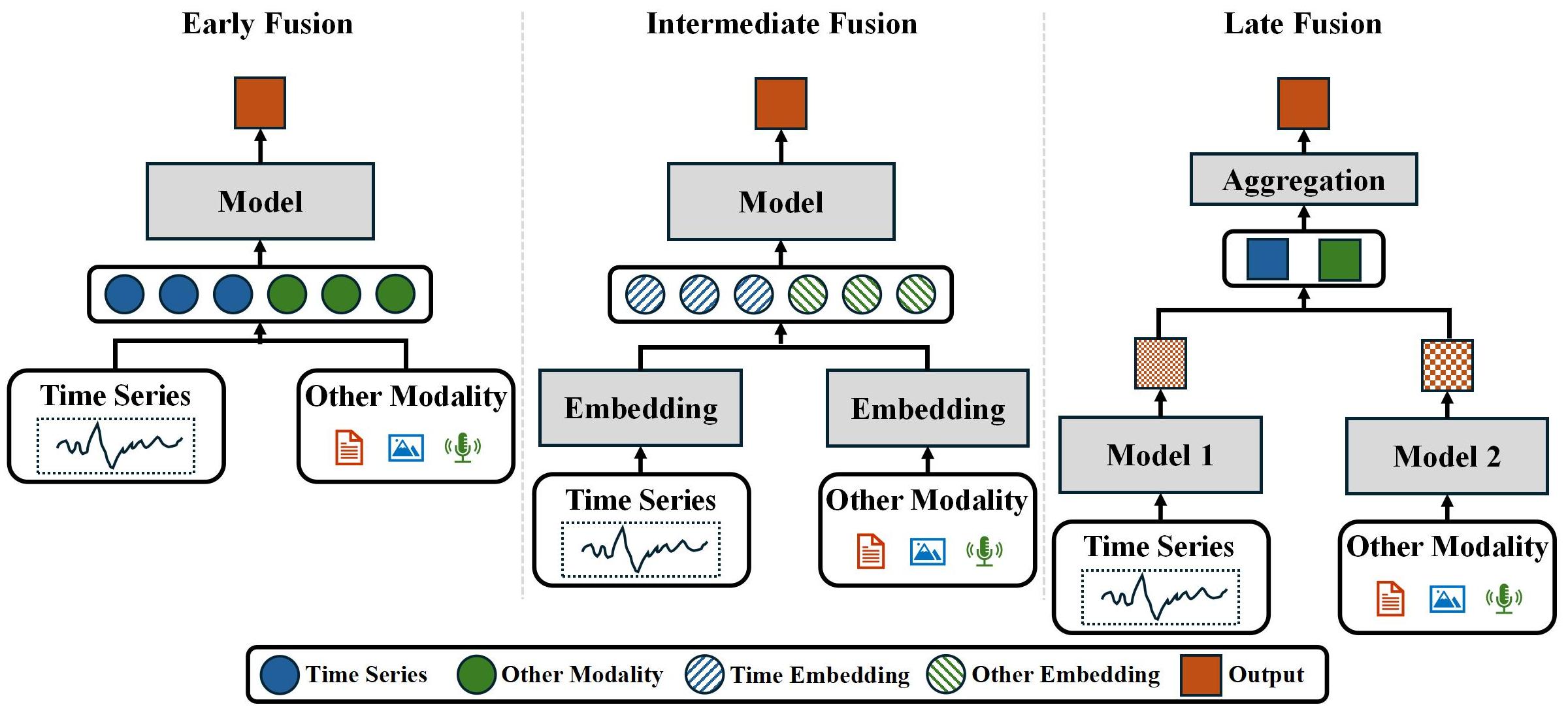}
    \caption{Taxonomy of Modality Fusion Solutions.}
    \label{pic:fusion}
\end{figure*}

\subsection{Time + Text}\label{sec:timeWithText}
We first introduce the types of integrated text, and then we introduce the key technology challenge: modality fusion.
\paragraph{\textbf{Integrated Text Types}}
We classify texts into two types: static texts and dynamic texts. Static texts, such as dataset descriptions~\cite{ming2023timellm}, offer static background and overall features such as data dimensions, global statistical information and task description~\cite{liu2024lstprompt}. In time-series classification tasks, the text can also convert traditional one-hot labels into textual labels~\cite{liu2024picture}. 

 Dynamic texts, such as real-time news, align closely with time series data. They include useful endogenous information from derived from data~\cite{jin2023time}; special events~\cite{wang2024chattime,xu2024beyond,liu2025autotimes} like holidays or weather events.
 Beyond providing the current context, dynamic text sequences can also provide the trajectory of context.  Most existing datasets of this type have three significant limitations: (1) Narrow data domains. Data characteristics and patterns vary between different domains, such as the periodicity of numerical data and the sparsity of textual data. However, current multimodal TS datasets~~\cite{dong2024fnspid,finance2,finance3,finance4,cao2023tempo} focus solely on stock prediction tasks in the financial domain, which are unable to represent the diverse data domains. (2) Coarse-grained modality alignment. Existing multimodal TS datasets only ensure that the text and numerical data come from the same domain, such as general stock news and the prices of one specific stock. Clearly, an abundance of irrelevant text diminishes the effectiveness of multimodal TSA. (3) Inherent data contamination.  Textual data often contains forecasting results, thus leads to biased evaluations. For example, the influenza outlook is regular in influenza reports. 

To this end, \citeauthor{liu2024timemmd} introduce Time-MMD, a multi-domain (nine) multimodal (numeric+text) time series dataset using real data. Specifically, it combines dynamic keyword search results and manually curated report sequences as aligned text series. Additionally, it disentangles descriptive and predictive text, for the data contamination issue.
In addition, \citeauthor{merrill2024language} and \citeauthor{williams2024context} construct synthetic datasets to evaluate context-guided TSF. 
Specifically, \citeauthor{merrill2024language} prompt GPT-4 to generate descriptions of environments that change over time alongside executable Python functions that generate corresponding time series. \citeauthor{williams2024context} present the ``Context is Key'' (CiK) benchmark, which consists of 71 manually designed forecasting tasks with manually created key text contexts such as \emph{``Suppose that there is a heat wave in city A..., resulting in excessive usage of only 5 times the usual electricity.''}. Time-IMM~\cite{chang2025time} studies irregular TS.

\paragraph{\textbf{Modality Fusion Solutions}}
The key challenge in integrating multiple modalities is modality fusion. Following the most common modality fusion taxonomy~\cite{atrey2010multimodal,boulahia2021early,stahlschmidt2022multimodal,zhao2024deep}, we classify existing works into three strategies: early, late, and intermediate fusion in Figure~\ref{pic:fusion}.

\underline{Early fusion} merges the original modalities at the input level. \citeauthor{williams2024context} validate the widespread effectiveness of direct prompting as an early fusion strategy in the constructed CiK benchmark.  \citeauthor{xinlei2024from} also adopt the early fusion strategy on dynamic text integration for text-guided TSF.  For classification tasks, InstructTime \cite{cheng2024advancing} and  HiTime \cite{tao2024hierarchical} also adopt early fusion to integrate domain description, prior knowledge and task descriptions. More details are provided in Appendix~\ref{app: fusion_e}. \underline{Late fusion} combines the separate outputs of each modality. MM-TSFlib~\cite{liu2024timemmd} first proposes a dual-path structure that combines LLMs and TSF models. Specifically, MM-TSFlib processes time series and textual data through separate specialized models before combining their outputs at the final layer.  \underline{Intermediate fusion} joins the representations of each modality.
 This type of works also adopts the dual-path structure similar to MM-TSFlib, but employs a time series encoder rather than a full time series model. GPT4MTS \cite{jia2024gpt4mts} combines time series and textual data by embedding each modality separately. before merging them at the embedding level. TGForecaster \cite{xu2024beyond} combines time series data with textual cues through cross-attention mechanisms.
 Texts as Time Series (TaTS) \cite{li2025language} treats time-series-paired texts as auxiliary variables and encode them into temporal representations and combine with numerical sequences at the embedding level.
 For classification tasks, DualTime~\cite{zhang2024dualtime} integrates time series and text data through dual adapters in a language model architecture. \citeauthor{xiongbeyond} use time pattern text prediction from LLMs as auxiliary information. LLM-TPF~\cite{panllm} extracts periodic features in the frequency domain and temporal trends from text. More details are provided in Appendix~\ref{app:fuse_i}.

\subsection{Time + Other Modalities}\label{sec:timeWithOther}
Compared to text, few studies focus on the combination of time series with other modalities. \citeauthor{girdhar2023imagebind} propose using vision to bind arbitrary modalities (including time series, represented by Inertial Measurement Unit (IMU) data, along with video, audio, image, and text), allowing the reuse of models with modality fusion capability. \citeauthor{ebrahimi2023lanistr} propose to integrate time series with text and tables for healthcare and retail scenarios.
\subsection{Domain-Specific Time-Series Works}\label{sec:timeWithDomain}
\paragraph{Financial Time-Series + X}
  Factors such as policies and social sentiment, typically represented in textual form, can significantly influence financial data and its dynamics~\cite{lee2009financial, xing2018natural}. Existing studies have demonstrated the benefit of effectively leveraging this textual information for better financial analysis with the numerical data. In particular, \citeauthor{emami2023modality} introduce a modality-aware transformer with intra-modal and inter-modal attention to improve stock price prediction. Few works \cite{kurisinkel2024text2timeseries,asgarov2023predicting,taylor2024natural} propose a language model that categorizes the positive or negative impact of sentiment on stock prices, further supporting stock price forecasting. 
\citeauthor{tavakoli2025multi} conduct an ablation study on CNN, RNN, and BERT encodings for mixed-modal inputs, incorporating self- and cross-attention mechanisms, and find that textual modality contributes even more than time series data for credit rating assessments. \citeauthor{rekabsaz2017volatility} integrate sentiment analysis, current volatility, GARCH, and sector-based features with an SVM using an RBF kernel for volatility prediction. 
Beyond finance analysis, \citeauthor{bamford2023multi} propose deep encoders for financial time series and textual data, enabling efficient financial data retrieval. \citeauthor{dong2024fnspid} introduce FNSPID, a financial domain-specific dataset that includes timestamps, articles, news, stock symbols, summaries, sentiment scores, URLs, multilingual content, and stock prices to support stock price prediction.

\paragraph{Spatial-Temporal Data + X}\label{subsubsec:spatialWithX}
There are also approaches that aim to benefit spatial-temporal time-series analysis by introducing extra modalities like texts or images. \citeauthor{anantharam2016understanding} propose to analyze city traffic dynamics with Twitter events. \citeauthor{fan2022urban} build a settlement classification model by inputting remote sensing images and sequences population density data. \citeauthor{lin2022spatial} propose to assist predicting Yahoo Finance prices by social media titles. \citeauthor{chen2024terra} establish a worldwide spatial-temporal dataset by combining climate series with geographical images and llm-derived climate descriptions. \citeauthor{feng2024citygpt} tune a language model by an originally established dataset to handle complex urban tasks involving images, trajectories, and texts. \citeauthor{lin2023mmst} encode both sensing images and short-term meteorological series jointly to help crop growth prediction. \citeauthor{zhang2024bjtt} establish a benchmark by aligning traffic records with traffic system descriptions. \citeauthor{han2024event} include extra event description texts to assist traffic forecasting. \citeauthor{jiang2024mobile} learn the effect of social media posts in predicting mobility statics in disaster including typhoon and COVID. \citeauthor{kosugi2022traffic} further consider toll and route search data. 

\paragraph{Medical Time Series + X} \label{Med:1}
In the medical domain, time-series analysis is often combined with text, image, and tabular data. The most widely used datasets are MIMIC  (Time-Series + Text + Image + Tabular) and  PTB-XL (Time-Series + Text). MIMIC-Based Datasets, with MIMIC-IV~\cite{johnson2020mimic} and MIMIC-III~\cite{johnson2016mimic}, containing electronic health records (EHRs), physiological time-series (vitals, lab results), and free-text clinical notes, supporting tasks like mortality prediction, ICU event forecasting, and disease progression. PTB-XL~\cite{wagner2020ptb} provides electrocardiogram (ECG) waveforms annotated with medical reports, enabling multimodal research that pairs time-series ECG signals with textual diagnosis descriptions for clinical classification tasks.  Image integration is widely studied, particularly in radiology, where physiological signals like electrocardiograms (ECGs) and patient vital signs are analyzed with medical images like X-rays, magnetic resonance imaging (MRI), and computed tomography (CT) scans to enhance diagnostic accuracy and clinical predictions~\cite{wang2024multimodal,yao2024addressing,hayat2022medfuse}. Time-series and tabular data fusion focuses on structured EHR data, combining static demographic, genetic, or laboratory test results with dynamic patient monitoring signals for predictive modeling~\cite{soenksen2022integrated, zhang2023improving,kyung2024towards}. Each modality combination serves specific medical tasks as detailed in Appendix~\ref{app:med+x}. Besides, a large number of multimodal studies have emerged on sequence structure data such as proteins~\cite{xiao2025protein}, RNA~\cite{xiao2024rna} and neuro signal~. But these works are beyond the scope of this time-series focused paper.

\section{Time2X and X2Time: Cross-Modality Interaction for Advanced TSA}\label{sec:time2X}
Time series modality lacks the natural interpretability compared with human-readable modalities such as text and images. Through cross-modality interaction between time series and human-readable modalities, existing Time2X and X2Time studies are revolutionizing the TSA paradigm, making it easier and more flexible for humans to use and understand.   We will introduce existing studies following the same order: first general TSA, then domain-specific TSA, with text modality followed by other modalities.
\subsection{Text 2 Time}\label{sec:text2Time}
\subsubsection{\textbf{Text2Time Retrieval}}
The Text2Time retrieval task retrieves time series data using text descriptions as queries, thereby largely reducing the difficulty and enhancing the flexibility of query design compared to traditional time-series query~\cite{assent2008ts,al2019time}. Existing work can be further categorized into two types: 

(1) Local retrieval, also known as feature understanding, focuses on detecting whether the feature type specified in the query exists in the time series. These works provide benefits through zero-shot generality and convenient verbalized interaction compared to supervised classification models. \citeauthor{fons2024evaluating} propose a comprehensive feature taxonomy and benchmark to evaluate whether LLMs can directly perform time series understanding tasks. Their results demonstrate that although LLMs at that time, such as GPT-4, perform well in identifying basic features such as trend and seasonality, they struggle to detect more complex aspects like structural breaks and volatility. Furthermore, \citeauthor{caitimeseriesexam} consider more recent LLMs, such as GPT-4o, and improve the evaluation procedure. 

(2) Most recently, \citeauthor{ito2024clasp} further study global retrieval, using overall textual descriptions to retrieve time series data, such as \emph{``It is a periodic signal. Rises at the beginning. Decreases at the end. Rises at the beginning.''} To this end, \citeauthor{ito2024clasp} propose a foundation retrieval model named CLaSP. \citeauthor{chen2025tracegroundingtimeseries} further study retrieval with additional context, such as event report. More details are provided in Appendix \ref{app:re}.
\subsubsection{\textbf{Text2Time Generation}}
Text2Time generation uses textual descriptions to guide and control the generation of time series data. Text2Time generation is typically used for synthesizing (large-scale) aligned text-time series datasets~\cite{merrill2024language,xie2024chatts}, or fulfilling instance-level requirements, such as generating ECG patterns that match a specific patient's disease condition~\cite{lai2025diffusets}.

\citeauthor{merrill2024language} first propose converting text constraints into code using LLMs, and then generating time series data through this code. \citeauthor{xie2024chatts} construct an ``attribute pool'' and corresponding ``generation rule pool'', such as mathematical functions, to make generation more precise and controllable. Most recently, \citeauthor{li2025bridge} propose BRIDGE which integrates a hybrid prompt mechanism combining semantic prototypes with textual descriptions in a diffusion model framework, thus enabling both domain-level guidance and instance-specific control.  VerbalTS~\cite{guverbalts} employs a multi-focal alignment and generation framework. \citeauthor{rousseau2025forging} further study multivariate time-series generation.
\subsection{Time 2 Text}\label{sec:time2Text}

\subsubsection{\textbf{Time2Text Captioning}}
The time series captioning task, also known as annotation, description, narration and summarization, involves generating natural language descriptions for a given time series, either comprehensively describing the data or briefly articulating key patterns with fluent expressions. Early research\cite{sharma2021t,li2023repr2seq} mostly focuses on introducing neural networks. \citeauthor{sharma2021t} propose a domain-independent neural framework for time series narration and demonstrate that this neural model can generate narratives with vocabulary diversity rather than using predefined templates. \citeauthor{spreafico2020neural} further propose a neural model for time-series line charts captioning. \citeauthor{li2023repr2seq} propose Repr2Se which uses TS2Vec~\cite{yue2022ts2vec} to extract core information from time series, and generates textual content through a neural network. 

Based on the recent rapid development of LLMs, \citeauthor{zhang2023insight} propose to model time series in text form and directly prompt GPT-4 to generate coherent descriptions. \citeauthor{dohi2024domain} point out that directly using LLMs for time series captioning tasks often leads to lower accuracy. They further train a contrastive learning-based model based on their constructed  dataset. \citeauthor{lin2024decoding} propose a multi-agent (general and domain-specific) annotation agent system for automatically generating both general and domain-specific time series captions. Most recently, \citeauthor{trabelsi2025time} propose TSLM which consumes a joint representation of text and embedding for a time series, and outputs a textual description of a time series as caption.
\subsubsection{\textbf{Time2Text Explanation}}
These works use natural language to explain black-box time series models. \citeauthor{liu2024largeAD} and \citeauthor{jiang2025explainable} propose intrinsic explanation methods for anomaly detection and event prediction correspondingly. \citeauthor{aksu2024xforecast} evaluate post-hoc language explanations of LLMs for TSF.

\subsection{Time 2 Text + Text 2 Time} \label{sec:timePlusText}

These works focus on multimodal multitask time series models, specifically handling text+time inputs and outputs, characterized by general time-series question \& answering (QA). ChatTime~\cite{wang2024chattime} reinterprets time series as a foreign language by normalizing, discretizing, and expanding the vocabulary of a pre-trained language model, thereby enabling bimodal inputs and outputs for multiple TSA tasks including zero-shot unimodal or multimodal forecasting and time-series QA. ChatTS~\citeauthor{xie2024chatts} bridges the gap between time series data and natural language by aligning LLMs with synthetic data. ChatTS is also designed for multimodal multivariate input and output for general-purpose time series tasks. \citeauthor{kong2025time} propose Time-MQA to enable comprehensive time series analysis by unifying diverse tasks---including forecasting, imputation, anomaly detection, classification, and open-ended reasoning---within a single natural language question answering framework. Time-MQA is designed to accept text+time input and produce text (time-series included) outputs for general-purpose TSA tasks across various domains.
\citeauthor{kim2024multi} focus on the joint text and time series forecasting tasks and introduce the Hybrid Multi-Modal Forecaster. ITFormer~\cite{leiitformer} explores a multi-task model that handles bidirectional text–time series inputs and outputs, with a focus on general QA systems.
\subsection{Other Cross-Modality Works}\label{sec:time2Others}
\citeauthor{islam2024datanarrative} propose to integrate visual aids, such as highlighted bars and lines in charts, along with text to automatically generate data stories from time series. They proposed a framework which uses a Generator and an Evaluator to simulate the human process of planning and revising narratives for this time2image+text task.
\subsection{Domain-Specific Time-Series Works} \label{sec:time2Domain}
\subsubsection{MedTime 2 X and X 2 MedTime. } \label{Med:2}

 \textbf{MedTime 2 Text. }
This type of cross-modality work involves converting medical time-series data (e.g., ECG, ICU monitoring signals) into structured natural language reports for diagnostic and clinical interpretation. Specific application scenarios include: (1) ECG Report Generation generating cardiology reports from ECG time-series data~\cite{bleich2024automated,wan2024meit,tangelectrocardiogram,wan2024electrocardiogram,li2024frozen} (2) ICU/NICU Data Summarization converting multichannel ICU/NICU physiological time-series data into summaries~\cite{hunter2008using, gatt2009data,hunter2008summarising} (3) Discharge Summary Automation generating coherent discharge summaries from time-series hospital course records~\cite{hartman2023method}.
 \textbf{Text 2 MedTime. }
In the medical domain, due to privacy concerns and limited medical resources, existing work focuses on using clinical text reports to synthesize realistic physiological signals, especially ECG~\cite{alcaraz2023diffusion,chung2023text,lai2025diffusets}.  \textbf{MedTime 2 Text + Text 2 MedTime.}
This type of work focuses on building chat models for medical time series analysis. ECG-Chat~\cite{zhao2024ecg}, ECG-Chat~\cite{guo2024multimodal} and ECG Semantic Integrator~\cite{yuecg} focus on QA for ECG signals; BioSignal Copilot~\cite{liu2023biosignal} further covers QA for multiple biosignals including ECG, EEG, PPG, EMG. MedTsLLM~\cite{chan2024medtsllm} can perform semantic segmentation, boundary detection and anomaly detection in time series, and includes text prompts to incorporate patient-specific information for deeper analysis of various physiological signals. Additionally, \citeauthor{cosentinotowards} builds medical models for personal health.
\subsubsection{FinTime 2 X and X 2 FinTime. }
 \textbf{FinTime 2 Text.} This type of cross-modality work for finance domain applications mainly focuses on (1) Financial Report Generation, converting time series data combined with other data types such as tables into readable financial reports, incorporating statistical models, market analysis, and risk assessment metrics~\cite{liu2022long,liu2023neural} (2) Market Analysis and Sentiment Reports, correlating economic indicators, market sentiment, and time series data to generate market interpretation text~\cite{chen2024knowledge,li2023multimodal}.
 \textbf{FinTime 2 Text + Text 2 FinTime. } This type of work builds foundation finance chat models~\cite{bhatia2024fintral,xie2024open}.

\subsubsection{Spatial-Temporal Analysis}
 \textbf{Spatial-Temporal Data 2 Text.} Unlike general TSA tasks, there haven't been progress around generating other modalities regarding spatial-temporal time-series input, or vice versa. The closest progress is around the language-based analysis of sequential images and events, including satellite image series \cite{irvin2024teochat,yan2024urbanclip} and urban event trajectories \cite{li2024urbangpt}. At the same time, recent advances in adapting LLMs to time-series data, as aforementioned in Section \ref{sec:timeWithText} and Section \ref{subsubsec:spatialWithX}, is capable of handling input from other modalities or adapting models from other modalities, but still targets on scenarios with time-series in both input and outputs. Instead, we argue that a comprehensive generative model, capable of handling input (or output) with modalities except time-series, are necessary for the development of spatial-temporal tasks as they can bridge the two modalities better.
 \textbf{Spatial-Temporal Data 2 Image.} Considering the complexity of spatiotemporal data, some studies focus on visualization solutions~\cite{wang2017spatial,yang2019research}.

\section{Gaps And Outlooks}
\subsection{For Modality Reusing: Which Modality?} \label{sec:modalityreuse}
As detailed in Section~\ref{sec:timeAsx}, existing works have made numerous efforts to reuse foundation models from multiple modalities, such as reformulated as text (with or without training), image (line graphs, heatmaps, spectrograms), audio, and tables. Evidently, a major gap lies in how to select the optimal modality repurposing strategy for a given TSA scenario, including different tasks, datasets, and objectives. Existing works have preliminarily demonstrated that this gap is not trivial, as there might not be a universal champion. Specifically, \citeauthor{hoo2024tabular} demonstrate that directly reformulating as tables to reuse foundation tabular models can even outperform specialized foundation time-series models; \citeauthor{zhou2024can} show that direct line graph reformulation is superior to direct text and table reformulation for anomaly detection. In addition, some recent studies \citeauthor{tan2024language,hu2025context} are still arguing the actual effectiveness of reusing language modality for TSF.

To mitigate this fundamental gap for TimeAsX studies, we propose several possible future directions. First, there is a need for a comprehensive benchmark as a guidebook which compares different modality repurposing solutions across a broad set of TSA scenarios. Second, automated methods for modality reuse may be useful. \citeauthor{liu2024picture} primarily shows that LLMs can correctly select frequency-domain or spectral-domain visualizations based on given time-series task descriptions. However, how to automatically select from the numerous reusing strategies requires further study. Finally, combining different time-series modeling strategies~\cite{ruan2025vision} can be a more effective solution, as each modality offers a unique view.
\subsection{For Modality Integration: How to Handle Heterogeneous Modality Combinations}
\label{sec:modalIntegration}

As detailed in Section~\ref{sec:timeAsx}, we can clearly observe the heterogeneity of modality combinations: (1) For general time series analysis, the usual combination is Time+Text. (2) For different domains, the combinations vary. For example, in the medical domain, Time+Tabular is common, while in other domains it is rare. (3) Within a specific domain, different datasets usually have different modality combinations. For instance, in the medical domain. the MIMIC dataset~\cite{johnson2020mimic} includes Time+Text+Image+Tabular, while PTXL only includes Time+Text~\cite{wagner2020ptb}. (4) Finally, at the sample level within a dataset, there are missing modality issues~\cite{ebrahimi2023lanistr,johnson2023mimic}. 

Apparently, handling these heterogeneous modality combinations is essential for real-world applications. We propose several possible future directions. The first is the scalable modality fusion method~\cite{girdhar2023imagebind} that is capable of handling a variable number of input modalities. Most current research is still limited to the fusion of fixed numbers (usually two) of modalities, as detailed in Section~\ref{sec:timeAsx}. The second is the robust method~\cite{ebrahimi2023lanistr} for handling missing modalities. One feasible solution is imputation, although current work is still limited to imputation for numerical modalities~\citet{du2024tsi}. Another potential solution is recently proposed by \citeauthor{liu2025maestro} to alleviate this problem via sparse Mixture-of-Experts mechanism.
\subsection{For Cross-Modality Interaction: How to Generalize to Unseen Tasks?}
\label{sec:crossmodalInteract}

As detailed in Section~\ref{sec:timeAsx}, we observe that current cross-modality studies have introduced several new TSA tasks, such as cross-modal retrieval, generation, and general question and answering. These emerging works support human-readable modalities, particularly language, as queries or (and) responses, significantly improving the usability and interpretability of TSA. However, the introduction of modalities like language makes it challenging to account for all cross-modal TSA tasks during training or pre-training phase, including unexplored tasks like time series editing~\cite{jing2024towards} guided by text, and unavoidable diverse user expressions.

To bridge this gap, we expect future time-series works to incorporate reasoning capabilities for unseen tasks. Compared to memory-based approaches, reasoning has been widely demonstrated to offer stronger generalization abilities~\cite{wei2022chain,guo2025deepseek}. The study of reasoning about time series is still in its early stages. We summarize several explored experiences from existing time-series research:  (1) TS-Reasoner~\cite{ye2024beyond}: task decomposition with coding,  (2) TimerBed and VL-Time~\cite{liu2024picture}: visualization with few-shot in-context learning, (3) Rec4TS~\cite{liu2025evaluating}: system 1 with test-time reasoning enhancement, and (4) ChatTS~\cite{xie2024chatts}: supervised fine-tuning with diverse data
\section{Conclusion}
We present the first survey for the emerging MM4TSA field, aiming to comprehensively answer: "How Can Time-Series Analysis Benefit from Multiple Modalities?" In this survey, we systematically identify three beneficial approaches: (1) TimeAsX: Reusing foundation models from other modalities for efficient TSA; (2) Time+X: Multimodal extensions for enhanced TSA; and (3) Time2X and X2Time: Cross-modality interaction for advanced TSA. For each category, we group existing works by modality type, introduce specific-domain applications, and discuss fundamental gaps with potential solutions. 
Our survey highlights the underlying trends of going beyond isolated time-series modalities studies, and could inspire
more innovative works within the field of MM4TSA. 
\section*{Acknowledgements}
This paper was supported in part by the NSF (Expeditions CCF-1918770, CAREER IIS-2028586, Medium IIS-1955883, Medium IIS-2106961, Medium IIS-2403240, PIPP CCF-2200269), CDC MInD program, Meta faculty gift, Dolby research gift and funds/computing resources from Georgia Tech and GTRI.

\bibliographystyle{ACM-Reference-Format}
\clearpage
\bibliography{references}
\appendix
\clearpage
\newpage
\section*{Appendix}
\section{More Detials of Background and Taxonomy}\label{app:back}
\subsection{Unimodal Time-Series Analysis Tasks}

Given an input multivariate time-series $X_{1:T} = \{x_1, x_2, \dots, x_T\}, x_i \in \mathcal{R}^m$, the corresponding tasks can be categorized into:

\noindent \textbf{Forecasting}, where the task objective is to predict the future part of $X$ given a certain look-back window. That is, $X_{1:t} \rightarrow X_{t:t+h}$, where $t$ is the look-back window length and $h$ is the forecasting horizon. 

\noindent \textbf{Imputation}, where the task objective is to accomplish the missing part of $X$. Given a mask $M_{1:T} = \{m_1, m_2, \dots, m_T\}, m_i \in \{0, 1\}$ noting if certain variable $x_i \in \mathbf{X}$ is missing, an imputation task can be formulated as $X_{1:T} \cdot M_{1:T}^T \rightarrow \mathbf{X}_{1:T}$.

\noindent \textbf{Anomaly Detection}, where the task objective is to detect possible anomalies inside $X$. Given anomaly labels $Y_{1:T} = \{\mathbf{y}_1, dots, \mathbf{y}_T\}, \mathbf{y}_i \in \{0, 1\}$ noting if certain variable $x_i \in \mathbf{X}$ belong to anomaly or not, an anomaly detection task can be formulated as $\mathbf{X}_{1:T} \rightarrow Y_{1:T}$.

\noindent \textbf{Classification}, where the task objective is to categorize $X$ into one of the classification labels. Given potential labels $\mathbf{Y} = \{\mathbf{y}_1, \mathbf{y}_2, \dots, \mathbf{y}_N\}$, a classification task can be formluated as $\mathbf{X}_{1:T} \rightarrow \mathbf{y}(X_{1:T}),\ \mathbf{y}(X_{1:T}) \in \mathbf{Y}$.

\subsection{Multimodal Time-Series Analysis}

In real-world time-series analysis, numerical time series often coexist with other data modalities (textual reports, images, categorical events, etc.). Tasks with such coexistence still fall into the four aforementioned categories but will involve input, output, or models from modalities other than numerical series. Depending on how such coexistence happens, we categorize multimodal time-series analysis into:

\noindent \textbf{Time with X}, where the inputs are coupled with other modalities aside original $X$. Given the observed modalities \( \{\mathbf{x}^{(m)}_t: m=1,\dots,M\} \) at time \( t \), the expected outputs are generated by $\hat{\mathbf{y}} = f(\mathbf{x}^{(1)}_{1:t}, \dots, \mathbf{x}^{(M)}_{1:t})$.

\noindent \textbf{Time as X}, where the original inputs are treated as an alternative modality and inputted to models designed for the corresponding modality. Given the alternative modality $m$, the modality transition function $\theta_m$ that changes time-series to modality $m$, and a domain-specific model $f_m$ for $m$, such multimodal task can be formulated as $\hat{\mathbf{y}} = f_m(\theta_m(\mathbf{x_{1:T}}))$.

\noindent \textbf{Time to X}, where the outputs of the task involve data in another modality. That is, $\hat{\mathbf{y}} = f(x_{1:T}^{(t)}) = \hat{\mathbf{y}}^{(m)}, \ m \neq t$
\section{Details on Time-series As Image}\label{sec:timeAsImageapp}
Reformulating time series as images for better feature perception is a natural idea, similar to how humans perceive patterns, and has received long-term research attention~\cite{zekun2023time,luoxiao2024vitime,zhuang2024see,xu2025can,yang2024vitime,daswani2024plots,mouxiang2024visionts,dosovitskiy2020image,zeng2021deep,namura2024training,roth2022towards,zhiguang2015imaging,kim2024cafo,li2020forecasting,siru2025timevlm}. For example, just as images have spatial relations between adjacent pixels, data points at nearby time steps show temporal relations. Furthermore, multivariate time series with many variates and long contexts can be effectively visualized as two-dimensional images using various methods. In this section, we showcase different ways to reformulate time series as images in order to reuse foundation vision models, such as the visual masked autoencoder (MAE)~\cite{he2022masked}.

\subsubsection{Line-graphs}
Using line-plots of time-series is the most popular way to convert time-series to images.
~\citet{zekun2023time} represented time-series a simple line-graph plots for univariate and multivariate time-series (stacked on top of each other).
They use a pre-trained swim transformer~\cite{liu2021swin} architecture to encode the input time-series
and a fine-tuned prediction head to perform time-series classification on human activity and healthcare benchmarks. They note that using line-graph allows modeling irregularly sampled time-series since the plot can adapt to missing values by interpolation on the line plot.
ViTime~\cite{luoxiao2024vitime} instead used a compressed image representation by mapping the time-series into binary pixels that indicate the time-series value at each time-step.

\citet{zhuang2024see} use line-graphs as images for few-shot anomaly detection using VLMs. 
They use a multi-step prompting method, first showing a set of normal
images to the model. Then they prompt the model to detect anomaly from a give time-series image.
They use multiple prompts to narrow down the candidate anomaly time-steps as well as type of anomaly as final candidate answer.
Both these methods show superior performance to state-of-the-art pre-trained foundational time-series models.
\citet{xu2025can} similarly use large multimodal VLMs to identify time series with anomalies. They upload an image containing multiple time-series in the form of a grid and prompt the VLM to choose which of the images are anomalous.

\citet{daswani2024plots} used a multimodal approach with line plots of time-series for wide range of time-series understanding tasks relevant to health related tasks.
They use in-context learning approach and used popular multimodal models such as GPT4o~\cite{hurst2024gpt} and Gemini Pro~\cite{team2024gemini}.
They showed that using a multimodal approach outperformed text-only, time-series-only alternatives of prompting for tasks such as activity recognition, fall detection and fitness readiness that require understanding patterns from multiple health-related and behavioral time-series.

\subsubsection{Heatmaps}
Heatmaps visualize time series in a 2D space using colors to represent magnitudes.
They are specifically useful for modeling long time series and multivariate time series.
For longer time series, VisionTS~\cite{mouxiang2024visionts} segments an input time series into patches and stacks the patches over each other.
They plot this matrix into a heatmap and feed it into Vision Transformer (ViT) ~\cite{dosovitskiy2020image} and pre-train it using masked auto-encoding and forecasting tasks. This approach provides state-of-the-art performance on zero-shot forecasting tasks across a wide range of benchmarks.
~\citet{zeng2021deep} uses heatmaps for multivariate time-series.
They visualize each of the time-series in grids and patch them together to form the output image.
They use a video generation model and input sequence of time-series images at different
consecutive windows to generate future forecasts in the generated frames.

\subsubsection{Spectrogram}
Time-series can be decomposed to spectrum of frequencies and represented as a spectrogram.
ITF-TAD~\citet{namura2024training} convert time-series into wavelet transform representation <cite>
and represent them as a spectrogram. When dealing with multivariate time-series, they use
PCA on frequencies to further reduce the number of channels for a compressed representation.
They use PatchCore~\cite{roth2022towards} as the vision foundational model to detect anomalies
in the time-series.
~\citet{zeng2023pixels} similarly used wavelet transforms with ViT for forecasting tasks.
They showed that using the spectrogram based representation is significantly more effective than line-plots.

\subsubsection{Other methods}
 \citet{zhiguang2015imaging} use Gramian Angular Fields (GAF)~\cite{campanharo2011duality}.
GAF converts time-series values into polar coordinates, where the angle is represented by trigonometric transformation of the time-series value and radius is represented from the time-stamp.
This representation can help visualize long and short term dependencies better by observing the diagonals of the image.
Recurrence plots (RP)~\citet{eckmann1995recurrence} are another way to capture periodic patterns in time-series by plotting time series with different frequencies in each row of the image.
~\citet{kim2024cafo} use RP representation and pre-trained vision models, including ViT, MLP-Mixer, and Resnet
for time-series classification.
~\citet{li2020forecasting} similarly apply it for forecasting.
Time-VLM~\cite{siru2025timevlm} is a multimodal technique which used fused representations from time-series, vision and text representations for the time-series for zero-shot and few-shot forecasting.
For image representation, they combine information from Fourier coefficients, cosine and sine periodicity into a heatmap which is fed into a VLM encoder.
The text information included task background information, basic statistics of the input time-series and description of the image.

\section{Details of Finance Time Series As Text}\label{Details of Finance Time Series As Image}
For instance, \citet{sezer2019financial} propose CNN-BI, which visualizes financial time series as bar images, enabling a model to learn buy-and-sell strategies for optimized trading decisions. \citet{pei2024stock} introduce SMSFR-CNN, which separately converts historical opening, highest, lowest, and closing prices, along with turnover rates (OHLCT), into images for accurate stock price prediction. \citet{zeng2023pixels} leverage time-series visualization with a Vision Transformer (ViT)~\cite{dosovitskiy2020image}, demonstrating improved forecasting accuracy compared to numerical forecasting models. \citet{du2020image} apply a wavelet transform to visualize stock prices for pattern extraction and denoising, followed by a CNN for stock trend analysis. \citet{chen2020encoding} propose GAF-CNN, which encodes candlestick charts using the Gramian Angular Field (GAF) representation and employs CNNs to classify eight critical candlestick patterns. \citet{hu2024research} utilize financial time-series visualizations to extract high-order features, enhancing the accuracy of risk prediction. Meanwhile, \citet{li2020forecasting} and \citet{wang2015imaging} explore general methods for transforming time series into visual formats, demonstrating improvements in forecasting, classification, and imputation on financial datasets. The key advantage of reframing financial time series as images lies in harnessing the power of vision models without introducing additional information beyond the original financial time series.  
\section{Details of Finance Time Series As Text}\label{Details of Finance Time Series As Text}
For example, \citet{yu2023temporal} prompt LLMs with financial time series and textual financial instructions, enabling the models to predict stock trend changes. Details of works within this line are provided in Appendix \ref{Details of Finance Time Series As Text}. Similarly, \citet{xie2023wall} incorporate financial time series and Twitter tweets as inputs, allowing LLMs to predict stock movements by integrating social sentiment. \citet{zhang2024multimodal} treat LLMs as agents, feeding them financial time-series data, images, and textual instructions to enhance stock return predictions. \citet{gan2024mme} propose MME-Finance that fine-tunes LLMs with financial time series represented as charts, alongside tables and documents, enabling LLMs to perform financial question answering (Q\&A). \citet{xie2024open} extend LLaMA with multimodal capabilities using financial time series, textual tokens, tables, and images, resulting in Open-FinLLMs that allow applications in financial sentiment analysis, classification, misinformation detection, credit scoring, and more. Beyond domain-specific approaches, general methodologies~\cite{gruver2023large, jin2023time, liu2024lstprompt} have also demonstrated success in improving forecasting accuracy on financial datasets by reframing financial time series as inputs suitable for LLMs. 

\section{Details of Finance Time Series As Graphs}\label{Details of Finance Time Series As Graphs}
This approach reformulates the identification of temporal correlations, such as relationships between stocks, as a temporal graph link prediction task, providing a novel perspective on financial time-series analysis, as well as \citet{foroutan2024deep}, that converts crude oil, gold, and silver markets time series to graph and captures the intricate interplay of spatial and temporal dependencies within crude oil and precious metals markets. \citet{cheng2022financial} propose MAGNN that converts financial time series to a heterogeneous graph network where the time series sources as nodes and relations in our financial knowledge graph as edges for finanace time series prediction.
\section{Details of Spatial-Temporal As Text}\label{Details of Spatial-Temporal As Text}
STG-LLM \cite{liu2024can} introduces two tokenizers seperately for time-series and textual prompts and then trains the model jointly. TrafficGPT \cite{zhang2024trafficgpt} combines LLMs with traffic foundation models by a unified pipeline capable of handling various traffic tasks. ST-LLM \cite{liu2024spatial} embeds input series with exogeneous spatial-temporal information and then feeds them into a partially frozen LLM. UniST \cite{yuan2024unist} introduces a learnable embedding prefix to adapt frozen LLM to forecast urban statics. These methods use LLM as part of the forecasting model and keep using original time-series as input and output, making them comparable to traditional time-series models.
\section{Details of Spatial-Temporal As Image}\label{Details of Spatial-Temporal As Image}
\citet{yao2018deep} propose to forecast taxi demand by combining temporal view, spatial view, and semantic view, where spatial view is encoded by convolutional neural networks (CNNs). \citet{li2022deep} use 3d convolutional networks to encode traffic trajectories and urban spatial correlations efficiently. \citet{bao2022storm} introduce generator-discriminator structure to predict human mobility responses in COVID. \citet{jiang2023learning} advance traffic forecasting by jointly encoding inputs by both graph neural networks and vision convolutional networks. \citet{tang2024vmrnn} combine vision Mamba with LSTM and produce a strong model capable of predicting spatial taxi demand. These methods are strong in encoding spatial information. Similar to LLM applications, they do not involve extra input information, but treat the original inputs in an alternative manner. Besides, neural signals ~\cite{wang2024exploring,wang2023extraction,li2024multi} can also be regarded as spatial-temporal data.

\section{More Details of Early Fusion} \label{app: fusion_e}
Specifically, \citeauthor{xinlei2024from} propose LLM-based multi-agent solutions to selectively integrate relevant news with time series data at the input stage before feeding both into a fine-tuned LLM. The model achieves alignment between news and time series through an iterative reasoning process where agents evaluate which news events meaningfully correlate with time series fluctuations, continually refining their selection logic based on validation results. 

Besides, for classification tasks, InstructTime \cite{cheng2024advancing}  integrates task instructions with raw time series data using a pre-trained LLM, reframing classification as a multimodal understanding task. Instead of classification to one-hot labels, InstructTime generates textual labels, improving comparability and cross-domain transferability. Specifically, InstructTime consists of three components: (a) a discretization module that converts time series into tokens, (b) an alignment layer that bridges the modality gap, and (c) a pre-training stage that enhances cross-domain generalization. \citeauthor{tao2024hierarchical} further propose a hierarchical solution, HiTime \cite{tao2024hierarchical}, for better dynamic temporal information modeling and modality alignment. HiTime employs a hierarchical feature encoder to capture diverse aspects of time series data using both data-specific and task-specific embeddings. To facilitate alignment, HiTime introduces a dual-view contrastive alignment module. Finally, with aligned embeddings as input, HiTime introduces a hybrid prompting strategy, i.e., including domain description, prior knowledge and tasks description, to prompt fine-tuned LLMs for time-series classifications.

\section{A More Detailed Introduction for Intermediate Fusion Works.}\label{app:fuse_i} This type of works also adopts the dual-path structure similar to MM-TSFlib, but employs a time series encoder to enhance LLMs rather than a full time series model. GPT4MTS \cite{jia2024gpt4mts} combines time series and textual data by embedding each modality separately (time series through patching with reversible instance normalization and textual data via BERT) before merging them at the embedding level. The model treating textual embeddings as trainable soft prompts that are prepended to the time series embeddings, which are then processed through transformer blocks with frozen attention mechanisms. TGForecaster \cite{xu2024beyond} combines time series data with textual cues (news embeddings and channel descriptions) through cross-attention mechanisms. The model aligns text and time series modalities through the modality mixer layer, where text embeddings serve as queries to synchronize with forecast lengths, enabling effective integration of external causal information for more accurate time series forecasting. Dual-Forecaster \cite{wu2025dualforecaster} integrates both historical descriptive text and future predictive text with time series data through a two-stage modality interaction framework. The model aligns modalities through three key mechanisms: a contrastive learning loss between historical text and time series, a history-oriented cross-attention module that creates aligned embeddings between time series and historical text, and a future-oriented cross-attention module that incorporates predictive textual insights into the already-aligned time series representations. 
Texts as Time Series (TaTS) \cite{li2025language} treats time-series-paired texts as auxiliary variables by encoding them into temporal representations that are combined with numerical sequences at the embedding level. The model capture the "Chronological Textual Resonance" phenomenon, where text representations exhibit periodic patterns similar to their corresponding numerical time series. TaTs \cite{li2025language} also use the Time-MMD dataset \cite{liu2024timemmd} to validate the effectiveness of multimodal (+text) expansion of imputation tasks.

Besides, for classification tasks, DualTime \cite{zhang2024dualtime} utilizes sample-level text, such as clinical reports for electrocardiograms or physician descriptions for electroencephalograms. DualTime integrates time series and text data through dual adapters in a language model architecture. DualTime uses learnable adaptation tokens injected into intermediate layers of a pretrained language model for multimodal fusion, with both temporal-primary and textual-primary adapters operating simultaneously and sharing the LM backbone to encourage embedding alignment and efficient fine-tuning. Notably, due to the scarcity of sample-level datasets, DualTime, despite being proposed as a general method, could only be validated in medical scenarios. 

\section{More Details of Text2Time Retrieval} \label{app:re}
\citeauthor{caitimeseriesexam} adopt procedural question generation and iterative refinement. Their benchmark, named TimeSeriesExam, shows that state-of-the-art closed-source VLMs at time, especially GPT-4o, already perform well on simple zero-shot feature understanding tasks, but still have difficulty handling more complex tasks, such as identifying causal relationships. CLaSP\cite{ito2024clasp} learns a joint multimodal space for time series signals and their textual annotations using contrastive learning, thus enabling zero-shot cross-modal global retrieval.
\subsubsection{More Details of  Medical Time Series + X}\label{app:med+x}
Each modality combination serves specific medical tasks. Time-series and text integration supports tasks such as mortality prediction, hospital readmission forecasting, disease phenotyping, and intensive care unit (ICU) deterioration modeling, where unstructured clinical notes complement physiological data in patient risk assessment~\cite{lee2023learning,wang2024multimodal}. Time-series and imaging fusion is commonly used in radiology for disease diagnosis and longitudinal tracking, such as leveraging sequential patient vitals and chest X-rays for pneumonia detection, pulmonary condition monitoring, and cardiovascular disease assessment~\cite{wang2024multimodal,yao2024addressing,hayat2022medfuse}. Time-series and tabular data fusion facilitates structured health risk modeling, as seen in treatment response prediction, where static baseline patient features (e.g., genetics, demographics) interact with dynamic disease progression markers~\cite{soenksen2022integrated, zhang2023improving,kyung2024towards}.

\end{document}